\documentclass[runningheads]{llncs}
\usepackage[T1]{fontenc}
\usepackage{graphicx,verbatim}
 \usepackage{amsmath} 
 \usepackage{amssymb} 
 \usepackage{booktabs}
\usepackage{multirow}
\usepackage{array}
\usepackage{adjustbox}
\usepackage{cite}
\usepackage{todonotes}
\usepackage{orcidlink}
\title{Frequency-Hierarchical Active k-Space Sampling for Diagnostic MRI}
\titlerunning{Frequency-Hierarchical Active k-Space Sampling for Diagnostic MRI}

\author{Anonymous Authors}
\authorrunning{Anonymous}
\institute{Anonymous Institution}

\author{Ruru Xu\inst{1} \and
Kian Anvari Hamedani\inst{2,3} \and
Zhikai Yang \inst{4} \and
Ilkay Oksuz\inst{1}\orcidlink{0000-0001-6478-0534}}
\authorrunning{Ruru Xu et al.}
\institute{Computer Engineering Department, Istanbul Technical University, Istanbul, Turkey \and
Department of Medical Biophysics, University of Toronto, Toronto, Ontario, Canada \and
Physical Sciences, Sunnybrook Research Institute, Toronto, Ontario, Canada \and
Department of Biomedical Engineering and Health, KTH Royal Institute of Technology, Stockholm, Sweden\\
\email{xu21@itu.edu.tr}}

\begin{document}
\maketitle

\begin{abstract}
Active sampling for accelerated MRI must distribute a tight sampling budget across spatial frequencies that carry very different kinds of information. Low frequencies hold most of the anatomical context; high frequencies carry the fine details that drive pathology assessment. Existing active samplers either treat both regions identically or restrict the action space to entire Cartesian rows, which forces a poor compromise at high acceleration. We propose \textbf{HieraSample}, a task-driven framework built around this hierarchy. A cosine-annealed curriculum lowers the acceleration factor from $20\times$ to $4\times$ across 80 acquisition steps while keeping a fully-sampled low-frequency disk at every step; a Mamba-based policy then picks individual high-frequency coordinates from features extracted by dual disease and severity classifiers. The reward is the per-sample reduction in class-weighted cross-entropy after each action, so a positive reward corresponds directly to a more confident correct prediction. On the fastMRI+ knee benchmark, HieraSample matches the fully-sampled oracle on ACL diagnosis from $4\times$ to $10\times$ acceleration, and improves on a recent Cartesian baseline by as much as $20.4$ AUC points on ACL severity.

\end{abstract}

\section{Introduction}
\label{sec:introduction}

Shortening MRI acquisition without losing diagnostic accuracy is one of the practical bottlenecks of clinical imaging. The standard remedy is to undersample k-space and reconstruct, either with compressed sensing~\cite{lustig2007sparse} or with learned priors~\cite{aggarwal2018modl,sriram2020end}, sometimes optimizing the mask and the reconstructor jointly~\cite{bahadir2020deep}. Most of these pipelines target image quality (PSNR, SSIM) and leave the downstream clinical task aside. Given a fixed sampling budget, which k-space coordinates maximize downstream diagnostic accuracy?

Recent active-sampling papers cast acquisition as sequential decision making~\cite{du2025active,xu2026adaptive}. Two recurring issues stand out. The first is geometric. Action spaces are usually uniform across k-space (typically full Cartesian rows), which discards the fact that low and high frequencies carry qualitatively different information. At aggressive acceleration this forces a bad compromise between anatomical context and pathological detail. The second is architectural. A policy that issues tens of sequential decisions needs an architecture whose cost does not grow with rollout length, which is awkward for both CNNs and standard transformers.

We propose \textbf{HieraSample}. Three design choices anchor the framework. (i) A \emph{cosine-annealed curriculum} drives $R(t)$ from $20\times$ to $4\times$ over $T{=}80$ steps. The initial mask is fixed to the smallest disk around the DC component, so a fully-sampled low-frequency core is guaranteed before any policy decision; the policy operates only on the high-frequency complement. (ii) \emph{Mamba state-space blocks}~\cite{gu2023mamba} appear on both sides of the system, in a dual-task disease/severity classifier and in the sampling policy. Their linear-time selectivity keeps the 80-step rollout cheap. (iii) The \emph{reward} at each step is the per-sample reduction in class-weighted cross-entropy after the action, not the raw loss. A positive reward then maps directly to a more confident correct prediction. The classifier and policy use one shared normalization scheme; mismatched normalization between the two pipelines quietly broke the reward signal at intermediate rollout steps in our early experiments.

On the fastMRI+ knee benchmark, HieraSample closes most of the gap to fully-sampled diagnosis on both ACL and cartilage classification across $4\times$ to $20\times$. Compared with ASSDM~\cite{du2025active}, retrained under matched data conditions, the framework gains $+5.0$ AUC points at $20\times$ on ACL diagnosis and $+20.4$ AUC points at $10\times$ on ACL severity. Comparing against an action-space-only ablation (Ours-Cartesian, same Mamba classifier and reward, Cartesian-row policy) further isolates where the point-based hierarchical action space adds value, most visibly on cartilage diagnosis where the classifier still has meaningful headroom to the oracle.

\section{Related Work}
\label{sec:related}

\noindent\textbf{Accelerated reconstruction.} Compressed sensing~\cite{lustig2007sparse} and unrolled networks such as MoDL~\cite{aggarwal2018modl} and VarNet~\cite{sriram2020end} take the sampling pattern as given and learn a reconstructor. Transformer-based and physics-informed variants~\cite{korkmaz2022unsupervised,feng2021task,peng2022learning} share that assumption; recent benchmarks such as the CMRxRecon2024 challenge~\cite{wang2025towards} extend these efforts across modalities and sampling patterns.

\noindent\textbf{Learned and adaptive sampling.} LOUPE~\cite{bahadir2020deep} learns a probabilistic mask jointly with the reconstructor, but the result is a single fixed pattern. Reinforcement-learning samplers~\cite{xu2026adaptive} adapt the mask during acquisition while still targeting image quality. ASSDM~\cite{du2025active} targets diagnostic accuracy via sequential decisions but treats k-space uniformly through Cartesian-row actions.

\noindent\textbf{State-space models.} Mamba~\cite{gu2023mamba} and S4~\cite{gu2021efficiently} provide linear-time sequence modeling. Recent work applies them to medical segmentation~\cite{ma2024u,wang2024mamba} and classification~\cite{yang2024mambamil,yue2024medmamba,zhu2024vision}; we are not aware of prior use in active acquisition.

\section{Method}
\label{sec:method}

\begin{figure}[t]
  \centering
  \includegraphics[width=0.85\linewidth]{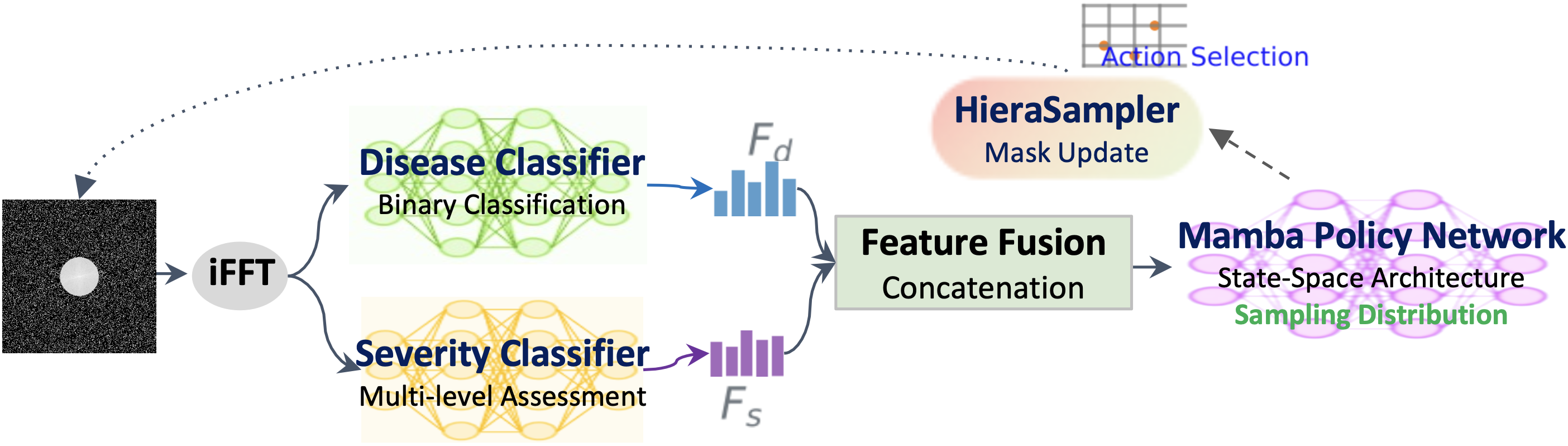}
  \caption{HieraSample framework. Mamba-based disease and severity classifiers extract features $\mathbf{F}_d,\mathbf{F}_s$ from the current zero-filled reconstruction. The Mamba policy fuses the two and emits sampling coordinates; HieraSampler combines them with the per-step budget $\Delta_t$ and updates the mask.}
  \label{fig:overall}
\end{figure}

\subsection{Problem formulation}
\label{sec:formulation}
Let $\mathbf{I}\in\mathbb{R}^{H\times W}$ be a ground-truth MRI slice and $\mathcal{F}\{\cdot\}$ the 2D Fourier transform. A binary mask $\mathbf{m}_t\in\{0,1\}^{H\times W}$ at step $t$ selects $|\mathbf{m}_t|$ of the $N{=}HW$ k-space samples; the acceleration is $R_t{=}N/|\mathbf{m}_t|$. The zero-filled reconstruction is $\mathbf{x}_t=\mathcal{F}^{-1}\{\mathbf{m}_t\odot\mathcal{F}\{\mathbf{I}\}\}$, with real and imaginary parts stacked into two channels and normalized per sample to have zero mean and unit standard deviation. Frozen disease ($G_d$) and severity ($G_s$) classifiers produce logits in $\mathbb{R}^2$. We seek each $\mathbf{m}_t$ such that the classifiers approach fully-sampled accuracy at the smallest possible $|\mathbf{m}_t|$.

\subsection{HieraSample acquisition curriculum}
\label{sec:hierasample}

\begin{figure}[t]
  \centering
  \includegraphics[width=0.92\linewidth]{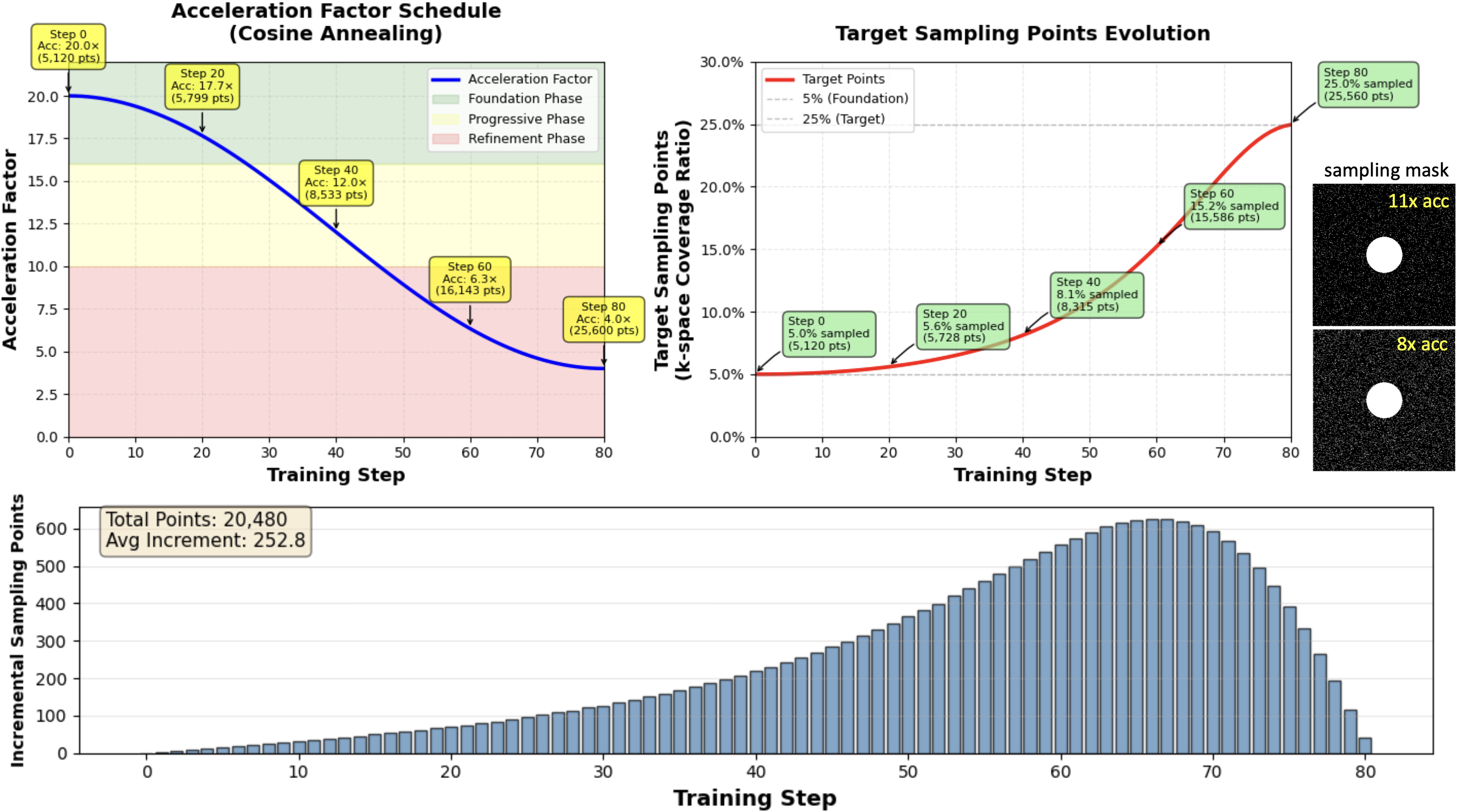}
  \caption{Progressive sampling under HieraSample. Left: cosine-annealed acceleration schedule from $20\times$ to $4\times$ over 80 steps. Middle: cumulative k-space coverage grows from $5\%$ to $25\%$. Right top: masks at three acceleration factors showing the preserved low-frequency disk plus scattered policy-chosen high-frequency points. Right bottom: per-step increment $\Delta_t$.}
  \label{fig:hierasample}
\end{figure}

\textbf{Cosine-annealed schedule.} The acceleration follows
\begin{equation}
  R(t) = R_{\max} - (R_{\max}-R_{\min})\cdot\tfrac{1}{2}\!\left(1-\cos\tfrac{\pi t}{T}\right),\quad R_{\max}{=}20,\;R_{\min}{=}4,
  \label{eq:cosine}
\end{equation}
with target $|\mathbf{m}_t|=\lfloor N/R(t)\rfloor$ and per-step increment $\Delta_t=|\mathbf{m}_t|-|\mathbf{m}_{t-1}|$. Increments are small near the endpoints and largest in the middle, where most of the diagnostic information is acquired (Fig.~\ref{fig:hierasample}).

\textbf{Low-frequency foundation.} $\mathbf{m}_0$ holds the $\lfloor N/R_{\max}\rfloor$ points nearest the k-space center, the smallest disk around DC. Anatomical low-frequency content is therefore present from $t{=}0$ and stays present at every subsequent step.

\textbf{Point-based action space.} At step $t$ the policy selects $\Delta_t$ new coordinates $\mathcal{S}_t$ from $\Omega\setminus\operatorname{supp}(\mathbf{m}_{t-1})$, and the mask updates as $\mathbf{m}_t=\mathbf{m}_{t-1}\cup\mathcal{S}_t$. Individual points, rather than entire rows or radial spokes, let the policy place the budget wherever diagnostic informativeness is highest.

\subsection{Mamba-based classifier}
\label{sec:classifier}

\begin{figure}[t]
  \centering
  \includegraphics[width=0.92\linewidth]{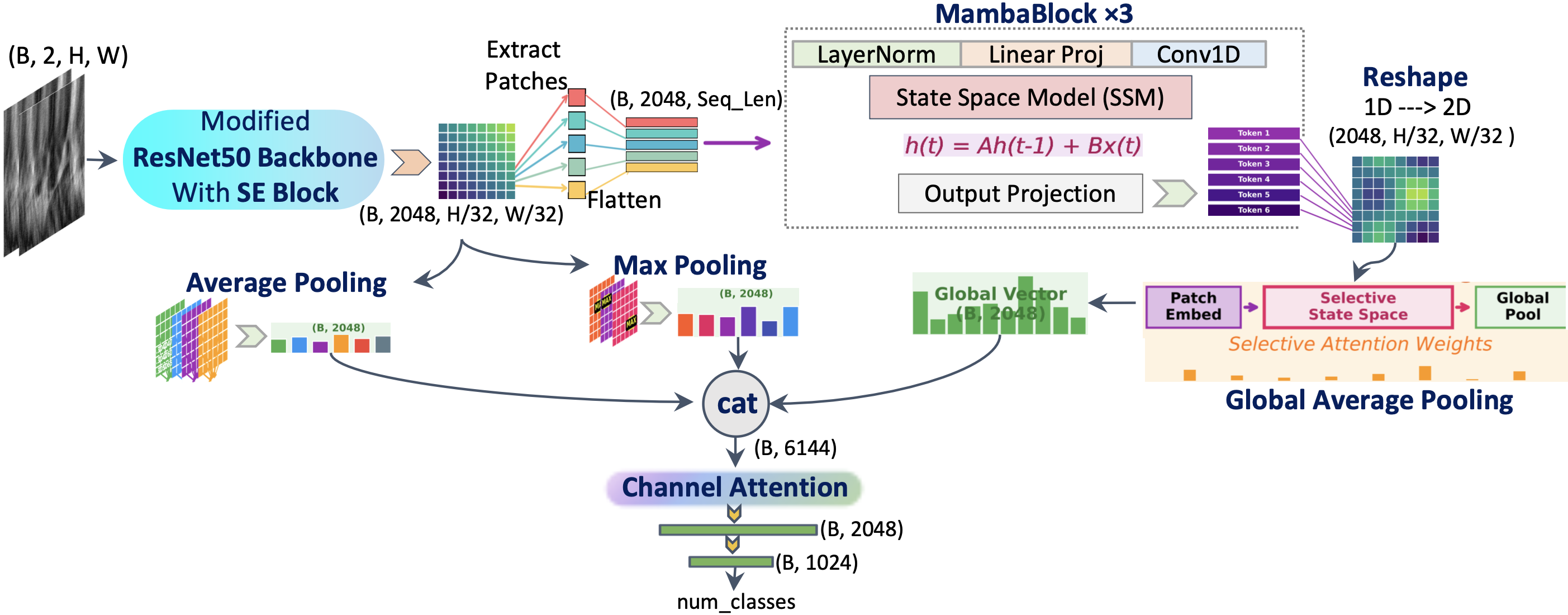}
  \caption{Mamba-based ResNet-50 classifier. The SE-ResNet-50 backbone yields a $(H/32){\times}(W/32){\times}2048$ feature map. Patches are linearized and refined by three Mamba blocks; the resulting global vector is concatenated with global average- and max-pooled backbone features and routed through a channel-attention MLP head.}
  \label{fig:classifier}
\end{figure}

Given a two-channel input $\mathbf{X}\in\mathbb{R}^{2\times H\times W}$, a ResNet-50 with squeeze-and-excitation blocks extracts a spatial feature map. The feature map is flattened into tokens and refined by three selective Mamba blocks~\cite{gu2023mamba}, each implementing
\begin{equation}
  \mathbf{h}_{k+1}=\overline{\mathbf{A}}_k\mathbf{h}_k+\overline{\mathbf{B}}_k\mathbf{x}_k,\qquad \mathbf{y}_k=\mathbf{C}_k\mathbf{h}_k+\mathbf{D}\mathbf{x}_k,
  \label{eq:mamba}
\end{equation}
with selectivity parameters predicted from $\mathbf{x}_k$. The global Mamba representation is concatenated with global average- and max-pooled backbone vectors, passed through a channel-attention block (Fig.~\ref{fig:classifier}), and reduced to two-class logits by an MLP head.

\textbf{Class-weighted CE on raw logits.} fastMRI+ is heavily imbalanced (1:28 for ACL, 1:13 for cartilage). We use cross-entropy on the raw logits with weights $\mathbf{w}=[1,\,n_{\mathrm{neg}}/n_{\mathrm{pos}}]$ taken from the training split, plus label smoothing $\alpha{=}0.1$. Softmax is applied only at evaluation; feeding softmax probabilities to PyTorch's \texttt{CrossEntropyLoss} would double-softmax the input and flatten the effective gradient.

\textbf{Normalization consistency.} Inputs are normalized with a single per-sample scalar mean and standard deviation, identical between classifier training and policy training. Without this match the classifier sees out-of-distribution zero-filled reconstructions at intermediate steps of the policy rollout, and the reward signal degenerates to noise.

\subsection{Mamba-based policy network}
\label{sec:policy}

\begin{figure}[t]
  \centering
  \includegraphics[width=0.92\linewidth]{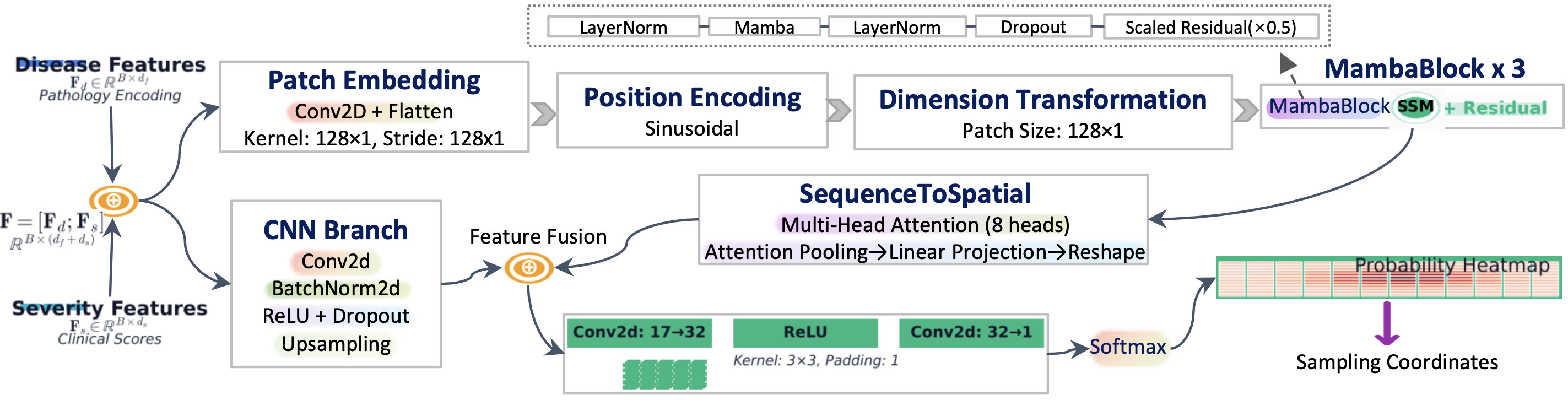}
  \caption{Policy network. Features $\mathbf{F}_d,\mathbf{F}_s$ are processed by parallel CNN and Mamba branches (rectangular $8{\times}16$ patch embedding plus three Mamba blocks); the fused output is a per-pixel sampling logit map. Coordinates are drawn by multinomial sampling without replacement at training and by top-$k$ at evaluation.}
  \label{fig:policy}
\end{figure}

The policy $\pi_\theta$ takes the concatenated classifier features $[\mathbf{F}_d;\mathbf{F}_s]$ and the current mask $\mathbf{m}_{t-1}$ (Fig.~\ref{fig:policy}). One branch is a CNN that captures local structure. The other branch applies a rectangular $8{\times}16$ patch embedding chosen to match the readout/phase-encoding anisotropy of k-space, yielding 320 tokens, then three Mamba blocks with $d_{\mathrm{model}}{=}128$, $d_{\mathrm{state}}{=}8$, expansion $1.5$, followed by an attention-pooled spatial decoder. The two branches are concatenated and a $1{\times}1$ convolution emits a per-pixel sampling logit map. Already-sampled positions are masked with $-\infty$, the remaining logits are softmaxed, and $\Delta_t$ coordinates are drawn without replacement (multinomial at training, top-$k$ at evaluation).

\subsection{Reward and policy gradient}
\label{sec:reward}
Let $\mathrm{CE}^{(d)}_t=\mathbf{w}_d^\top\mathrm{CE}(G_d(\mathbf{x}_t),y_d)$ denote the class-weighted cross-entropy of the disease classifier on $\mathbf{x}_t$, and $\mathrm{CE}^{(s)}_t$ the analogous quantity for severity, set to zero on disease-negative samples where the severity label is undefined. The per-task reward at step $t$ is the cross-entropy reduction the action produced:
\begin{equation}
  r^{(d)}_t=\mathrm{CE}^{(d)}_{t-1}-\mathrm{CE}^{(d)}_t,\qquad r^{(s)}_t=\mathrm{CE}^{(s)}_{t-1}-\mathrm{CE}^{(s)}_t.
  \label{eq:reward}
\end{equation}
Using the \emph{reduction} rather than the loss itself makes positive rewards correspond to diagnostic improvement. The two task rewards are combined with a cosine weight $\lambda_t=\tfrac{1}{2}(1-\cos\tfrac{\pi(t+1)}{T})$, $r_t=(1-\lambda_t)r^{(d)}_t+\lambda_t r^{(s)}_t$. Early steps emphasize disease detection; later steps emphasize severity grading. A per-batch baseline is subtracted from $r_t$ for variance reduction, and the policy is updated by REINFORCE with gradient norm clipped at $1.0$.

\section{Experiments}
\label{sec:experiments}

\subsection{Setup}
\textbf{Data.} We evaluate on the fastMRI+~\cite{zhao2022fastmri+} knee dataset, which annotates fastMRI~\cite{zbontar2018fastmri} acquisitions with pathology labels. ACL injury has $1{,}341$ pathology-positive slices against $37{,}829$ negatives; within the training split alone the positives split into Low ($617$) and High ($503$) severity grades, with $114$ and $107$ further positives appearing in validation and test respectively. Cartilage lesion has $2{,}763$ positives against $36{,}407$ negatives, graded as partial- vs full-thickness. We use the official $32{,}035$ / $5{,}294$ / $1{,}841$ train/val/test split for both tasks.

\textbf{Acceleration augmentation.} Each training slice is materialized at five acceleration factors $\{4,6,8,10,20\}\times$, giving a $5\times$ effective training set. Validation and test use a fixed acceleration per evaluation. The same $5\times$ augmentation is applied to the ASSDM baseline, which we retrain from scratch under our exact data conditions for a fair comparison.

\textbf{Implementation.} The classifier uses Adam ($\mathrm{lr}{=}4{\times}10^{-4}$, step decay $\gamma{=}0.97$ every 2 epochs), class-weighted CE with label smoothing $0.1$, up to $100$ epochs, with checkpoint selection on validation ROC AUC. The policy is trained for $50$ epochs (Adam, $\mathrm{lr}{=}10^{-4}$, $\gamma{=}0.5$ every 5 epochs, batch $50$) on top of frozen classifiers, $T{=}80$. All runs use a single NVIDIA A100 and a fixed seed. The frozen classifier has 274M parameters; the policy adds 53M trainable parameters.

\textbf{Baselines and metrics.} We structure the comparison into two parts. \textbf{ASSDM-Cartesian}~\cite{du2025active} reproduces the full published pipeline (its original classifier and Cartesian-row policy) retrained from scratch on our augmented data; this comparison therefore reflects the combined effect of classifier and policy. \textbf{Ours-Cartesian} and \textbf{Ours-hierarchical} share our Mamba classifier, the Mamba policy backbone, the cosine-annealed acquisition schedule (driving $R$ from $20\times$ to $4\times$ over 80 steps), the per-step budget $\Delta_t$, the low-frequency disk initialization $\mathbf{m}_0$, and the reward; they differ only in how each step's budget is spent (entire Cartesian rows for Ours-Cartesian versus individual high-frequency coordinates for Ours-hierarchical). Comparing these two therefore isolates the action-space contribution. We report balanced accuracy (BA at threshold $0.5$) and ROC AUC, with paired-bootstrap $95\%$ confidence intervals of $\pm 0.003$ AUC on the larger evaluation pools and $\pm 0.015$ AUC on the ACL severity test set (107 positives). The fully-sampled classifier is the oracle upper bound.

\subsection{Results}

\begin{table}[t]
\centering
\caption{ACL diagnosis and severity. Bold marks the best value in each column across undersampled methods. ROC AUC is reported alongside BA because the $1{:}28$ class imbalance makes a threshold of $0.5$ a poor proxy for ranking quality. Cells where an undersampled method exceeds the oracle by $\leq 0.001$ (ACL diagnosis $4\times$, ACL severity $4{\times}/6{\times}$) lie within the single-seed paired-bootstrap confidence interval and should be read as ``matches the oracle'' rather than ``exceeds the upper bound''.}
\label{tab:acl_results}
\setlength{\tabcolsep}{2.5pt}
\renewcommand{\arraystretch}{1.0}
\small
\resizebox{\textwidth}{!}{%
\begin{tabular}{@{}lcccccc|ccccc@{}}
\toprule
\multirow{2}{*}{\textbf{Method}} & \multicolumn{5}{c|}{\textbf{ACL Diagnosis}} & & \multicolumn{5}{c}{\textbf{ACL Severity}} \\
\cmidrule{2-6}\cmidrule{8-12}
 & 20$\times$ & 10$\times$ & 8$\times$ & 6$\times$ & 4$\times$ & & 20$\times$ & 10$\times$ & 8$\times$ & 6$\times$ & 4$\times$ \\
\midrule
\multicolumn{12}{c}{\textit{Balanced Accuracy} $\uparrow$} \\
\midrule
Fully Sampled (oracle)   & \multicolumn{5}{c|}{0.891}        & & \multicolumn{5}{c}{0.807} \\
ASSDM-Cartesian          & 0.750 & 0.812 & 0.817 & 0.827 & 0.833 & & 0.629 & 0.642 & 0.654 & 0.662 & 0.654 \\
Ours-Cartesian           & \textbf{0.877} & 0.879 & 0.881 & \textbf{0.887} & \textbf{0.886} & & \textbf{0.717} & \textbf{0.717} & 0.720 & 0.721 & 0.723 \\
Ours-hierarchical        & 0.874 & \textbf{0.886} & \textbf{0.886} & 0.886 & 0.885 & & 0.707 & 0.713 & \textbf{0.721} & \textbf{0.725} & \textbf{0.725} \\
\midrule
\multicolumn{12}{c}{\textit{ROC AUC} $\uparrow$} \\
\midrule
Fully Sampled (oracle)   & \multicolumn{5}{c|}{0.939}        & & \multicolumn{5}{c}{0.837} \\
ASSDM-Cartesian          & 0.875 & 0.887 & 0.887 & 0.895 & 0.900 & & 0.680 & 0.616 & 0.677 & 0.695 & 0.686 \\
Ours-Cartesian           & \textbf{0.928} & 0.930 & 0.923 & \textbf{0.939} & \textbf{0.940} & & 0.817 & \textbf{0.824} & \textbf{0.835} & 0.836 & \textbf{0.838} \\
Ours-hierarchical        & 0.925 & \textbf{0.939} & \textbf{0.939} & \textbf{0.939} & 0.938 & & \textbf{0.822} & 0.820 & 0.825 & \textbf{0.838} & 0.829 \\
\bottomrule
\end{tabular}}
\end{table}

\begin{table}[t]
\centering
\caption{Cartilage diagnosis and severity. Differences within $\pm 0.003$ are within the paired-bootstrap $95\%$ confidence interval and are not statistically significant.}
\label{tab:cart_results}
\setlength{\tabcolsep}{2.5pt}
\renewcommand{\arraystretch}{0.9}
\small
\resizebox{\textwidth}{!}{%
\begin{tabular}{@{}lcccccc|ccccc@{}}
\toprule
\multirow{2}{*}{\textbf{Method}} & \multicolumn{5}{c|}{\textbf{Cart.\ Diagnosis}} & & \multicolumn{5}{c}{\textbf{Cart.\ Severity}} \\
\cmidrule{2-6}\cmidrule{8-12}
 & 20$\times$ & 10$\times$ & 8$\times$ & 6$\times$ & 4$\times$ & & 20$\times$ & 10$\times$ & 8$\times$ & 6$\times$ & 4$\times$ \\
\midrule
\multicolumn{12}{c}{\textit{Balanced Accuracy} $\uparrow$} \\
\midrule
Fully Sampled (oracle)   & \multicolumn{5}{c|}{0.755}        & & \multicolumn{5}{c}{0.641} \\
ASSDM-Cartesian          & 0.701 & 0.702 & 0.704 & 0.719 & 0.721 & & 0.604 & 0.613 & 0.617 & 0.617 & 0.623 \\
Ours-Cartesian           & \textbf{0.709} & 0.705 & \textbf{0.711} & \textbf{0.719} & \textbf{0.733} & & \textbf{0.625} & \textbf{0.641} & \textbf{0.648} & 0.647 & \textbf{0.631} \\
Ours-hierarchical        & \textbf{0.709} & \textbf{0.709} & 0.709 & 0.718 & 0.732 & & \textbf{0.625} & \textbf{0.641} & \textbf{0.648} & \textbf{0.648} & \textbf{0.631} \\
\midrule
\multicolumn{12}{c}{\textit{ROC AUC} $\uparrow$} \\
\midrule
Fully Sampled (oracle)   & \multicolumn{5}{c|}{0.902}        & & \multicolumn{5}{c}{0.802} \\
ASSDM-Cartesian          & 0.842 & 0.852 & 0.855 & 0.856 & 0.861 & & 0.642 & 0.663 & 0.665 & 0.651 & 0.716 \\
Ours-Cartesian           & 0.851 & 0.856 & 0.861 & 0.867 & 0.869 & & \textbf{0.681} & 0.670 & \textbf{0.683} & \textbf{0.718} & \textbf{0.731} \\
Ours-hierarchical        & \textbf{0.866} & \textbf{0.865} & \textbf{0.867} & \textbf{0.870} & \textbf{0.870} & & 0.680 & \textbf{0.672} & 0.682 & \textbf{0.718} & \textbf{0.731} \\
\bottomrule
\end{tabular}}
\end{table}

\begin{figure}[t]
  \centering
  \includegraphics[width=\linewidth]{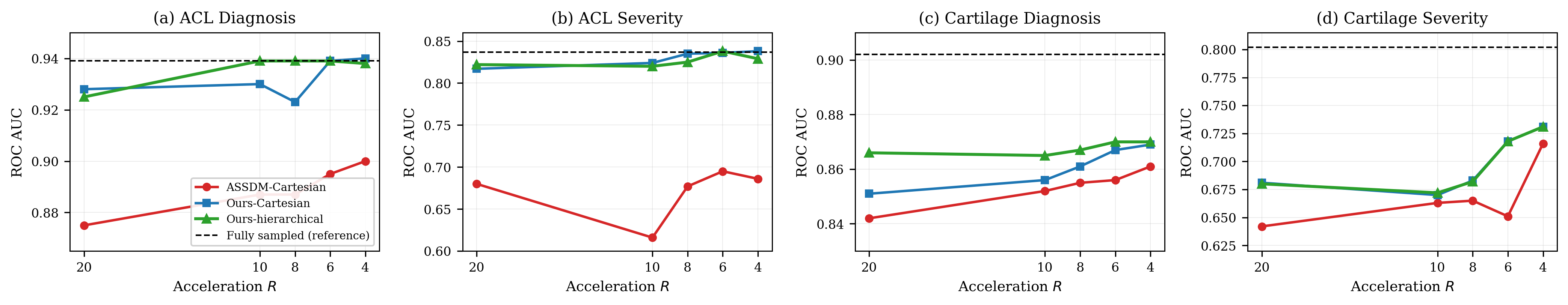}
  \caption{ROC AUC versus acceleration factor on the four fastMRI+ classification tasks; the dashed line is the fully-sampled reference. HieraSample (green) and Ours-Cartesian (blue) both close most of the gap to the reference; ASSDM (red) trails by margins exceeding $13$ AUC points on ACL severity at every acceleration factor. The two Mamba-based variants overlap on most tasks, with the hierarchical action space pulling ahead on cartilage diagnosis. Differences of order $0.001$ between an undersampled method and the reference (e.g.\ Ours-Cartesian at $4\times$ in panel (a)) lie within paired-bootstrap test-set variance under a single seed and should be read as ``matches the reference'' rather than ``exceeds the upper bound''.}
  \label{fig:auc_curves}
\end{figure}

\textbf{Improvements over the published baseline.} HieraSample's largest single-cell gain over ASSDM lands on ACL severity at $10\times$ acceleration: ROC AUC rises from $0.616$ to $0.820$, a $20.4$-point jump. Across the five accelerations, ASSDM's ACL-severity AUC stays in the $0.616$ to $0.695$ band while HieraSample stays at or above $0.820$, giving a pointwise gain of at least $14.2$ AUC points. ACL diagnosis follows the same shape but with a smaller gap, $+5.0$ AUC points at $20\times$ narrowing to $+3.8$ at $4\times$. Cartilage diagnosis sits in between ($+2.4$ at $20\times$, $+0.9$ at $4\times$).

\textbf{Approaching the fully-sampled oracle.} On ACL diagnosis Ours-hierarchical reaches the oracle's $0.939$ ROC AUC at $10\times$, $8\times$, and $6\times$, and comes within $0.001$ at $4\times$; the gap only opens to $0.014$ at $20\times$, where the sampling budget drops to $5\%$ of k-space. On ACL severity at $6\times$, $0.838$ matches the oracle of $0.837$ within a thousandth of a point. Cartilage diagnosis at $4\times$ retains a $0.032$ gap.

\textbf{Hierarchical vs.\ Cartesian under the same classifier.} On ACL diagnosis the two are close: BA values fall within $0.007$ across the five accelerations; Ours-hierarchical wins ROC AUC at $10\times$ by $+0.009$ and at $8\times$ by $+0.016$, while Ours-Cartesian wins by $+0.003$ at $20\times$ and $+0.002$ at $4\times$ (all within or near the $\pm 0.003$ bootstrap envelope). On cartilage diagnosis Ours-hierarchical wins at every acceleration, with gains of $+0.015, +0.009, +0.006, +0.003, +0.001$ from $20\times$ to $4\times$; the first three exceed the $\pm 0.003$ envelope and the trend decays monotonically. The two severity tasks overlap within the larger $\pm 0.015$ envelope. The pattern follows classifier headroom: cartilage diagnosis has a $0.032$ oracle gap at $4\times$ and shows hierarchical gains, while ACL diagnosis is already within $0.001$ of the oracle and leaves nothing for the action space to add.

\section{Conclusion}
\label{sec:conclusion}
HieraSample is a task-driven active MRI sampling framework that combines a cosine-annealed curriculum, a preserved low-frequency foundation, and a Mamba policy that places individual high-frequency coordinates. On the fastMRI+ knee benchmark it matches the fully-sampled oracle on ACL diagnosis from $4\times$ to $10\times$ acceleration and improves on the ASSDM baseline by up to $+20.4$ AUC points on ACL severity, at a rollout cost of $0.4$\,s per slice on an A100. The evaluation is restricted to single-coil single-anatomy knee data under a single random seed, and our action-space ablation isolates rows versus points but does not separately attribute the cosine schedule and low-frequency foundation; jointly fine-tuning the classifier on the policy's own mask distribution and disentangling the schedule and foundation are left to future work. 

\begin{credits}
\subsubsection{\ackname} 
This paper has been produced benefiting from the 2510 Program of The Scientific and Technological Research Council of Türkiye (project no: 124N419). The paper also benefited from Istanbul Technical University Scientific Research Projects (ITU BAP) funds, grant number 47653.
\end{credits}

\bibliographystyle{splncs04}

\end{document}